\documentclass[10pt,twocolumn,letterpaper]{article}

\usepackage[pagenumbers]{wacv}

\usepackage{multirow}          

\usepackage{tikz}
\usetikzlibrary{positioning, arrows.meta, calc, fit, backgrounds}
\definecolor{tcol}{RGB}{78,100,138}    
\definecolor{scol}{RGB}{20,135,125}    
\definecolor{lcol}{RGB}{222,118,86}    

\graphicspath{{figures/}}      

\newcommand{\R}{\mathbb{R}}                 
\newcommand{\teacher}{g_{\phi}}             
\newcommand{\student}{f_{\theta}}           
\newcommand{\stopgrad}{\operatorname{sg}}   

\newcommand{\myparagraph}[1]{\vspace{3pt}\noindent{\bf #1}}

\usepackage{bm}

\definecolor{wacvblue}{rgb}{0.21,0.49,0.74}
\usepackage[pagebackref,breaklinks,colorlinks,allcolors=wacvblue]{hyperref}

\def\wacvPaperID{1107}
\def\confName{WACV}
\def\confYear{2027}

\title{Cross4D-JEPA: Dense Cross-modal Correspondence Distillation \\ for 4D Point Cloud Representation Learning}

\author{
Trung Thanh Nguyen$^{1}$,
 Hai Nguyen-Truong$^{2}$, Tu Vo$^{3}$, Hoang M. Truong$^{4}$, and Tuan-Anh Vu$^{5}$\thanks{Corresponding author: {tuananh.vu@ucla.edu}.}
\\
$^{1}$Nagoya University, Japan \hspace{5pt}
$^{2}$Northeastern University, USA  \hspace{5pt}
$^{3}$KC Machine Learning Lab, Korea \\
$^{4}$University of Science, Vietnam National University Ho Chi Minh City, Vietnam \\
$^{5}$University of California, Los Angeles, USA
}

\begin{document}
\maketitle


\begin{abstract}
    Automatic understanding of dynamic 4D point clouds, the 3D-point sequences captured over time by depth sensors and LiDAR, is central to robotics and embodied perception.
    Yet annotating them densely is expensive, making self-supervised pretraining the natural route to transferable representations.
    Existing pretext tasks, however, are almost entirely intra-modal, and the few methods that transfer knowledge from 2D foundation models rely on a single global embedding per clip, discarding the rich per-patch semantics that these models compute.
    To address this gap, we propose \textbf{Cross4D-JEPA}, a teacher--student method that distills a frozen 2D foundation model, an image model DINOv2, or a video model V-JEPA\,2, into a 4D point encoder.
    The proposed method combines \textbf{(1)} a dense cross-modal correspondence that maps every 3D point to the teacher patch feature it projects to, and \textbf{(2)} a per-point objective that trains the student to match these features in latent space with no masking, negatives, or decoder.
    We evaluate Cross4D-JEPA on four benchmarks, MSR-Action3D, DeformingThings4D, NTU-RGB+D\,60, and HOI4D, against intra-modal and global cross-modal baselines.
    Experimental results show that, under a matched protocol, the proposed method consistently outperforms intra-modal and global cross-modal baselines across the four benchmarks and is competitive with heavier published 4D methods; further analysis attributes this gain primarily to the granularity of the correspondence rather than the teacher modality.
    Beyond recognition accuracy, the dense representation learned by Cross4D-JEPA transfers across domains, improves label efficiency, and improves full-label fine-tuning under the same training budget, while a $13\times$ smaller encoder matches a heavyweight pooling backbone.
\end{abstract}
    
\section{Introduction}
\label{sec:intro}

Automatic understanding of dynamic 4D point clouds, the sequences of 3D points captured over time by depth sensors, LiDAR, and multi-view rigs, is central to robotics and embodied perception~\cite{pstnet,p4transformer,pptr}, where recognizing actions, interactions, and deformations~\cite{msraction3d,nturgbd,hoi4d,deformingthings4d} underlies downstream decisions.
However, dense temporal labels are expensive to obtain, making self-supervised pretraining the natural route to transferable 4D representations~\cite{strl,mastpre,uni4d}.

Yet existing pretext tasks for 4D point clouds are almost entirely \emph{intra-modal}.
Spatio-temporal contrastive learning~\cite{strl,pointcmp,fourdcontrast}, masked structure prediction~\cite{mastpre,uni4d}, and complete-to-partial geometry distillation~\cite{c2p} all supervise the encoder from the point clouds alone, and never inject the appearance and object semantics that large 2D models already capture.
The one method that transfers from 2D video into a point encoder, CrossVideo~\cite{crossvideo}, contrastively matches a single global embedding per clip.
This signal is coarse, collapsing a structured 2D feature map into a single vector and discarding where each feature came from.

This is a missed opportunity, as the richest 2D foundation models are precisely those with spatially dense, per-patch features that a single global embedding collapses.
DINOv2~\cite{dinov2} produces a patch-level feature map whose local descriptors support dense correspondence and segmentation without any fine-tuning.
Static cross-modal distillation already exploits this property, as CrossJEPA~\cite{crossjepa} and Concerto~\cite{concerto} distill an image model into a \emph{static} point encoder by predicting target embeddings in latent space~\cite{ijepa}, the ``beyond-masking'' recipe that also underlies Point-JEPA~\cite{pointjepa}.
The open question is how to carry such knowledge into \emph{dynamic} point clouds, and whether to do so at the granularity of one embedding per clip, as in prior video-to-point transfer, or densely, per point.

To close this gap, we propose \textbf{Cross4D-JEPA}, a teacher--student method that distills a frozen 2D foundation model, an image model DINOv2, or a video model V-JEPA\,2~\cite{vjepa2}, into a geometry-only 4D point encoder.
The proposed method has two components.
First, it builds a dense cross-modal correspondence.
For each frame, we render the point cloud, record the per-pixel nearest-point index, and pull every teacher \emph{patch} feature back onto the 3D point it represents, giving an exact, occlusion-aware geometric correspondence.
Second, a per-point objective trains the student to predict, in the teacher's latent space, the teacher embedding for each point, without masking, negatives, or a decoder.
With the teacher frozen and its per-point targets cached once, pretraining incurs no additional teacher forward pass and stays within single-GPU hours.
The main contributions of this work are as follows:
\begin{itemize}
    \item We propose \textbf{Cross4D-JEPA}, a dense cross-modal correspondence distillation method that distills a frozen 2D foundation model into a 4D point encoder per point, rather than via a single global embedding per clip, without masking, negatives, or a decoder.
    \item We show through controlled analysis that the transfer of Cross4D-JEPA is driven by correspondence {granularity}, not teacher {modality}, as the image teacher DINOv2~\cite{dinov2} and the video teacher V-JEPA\,2~\cite{vjepa2} are statistically tied under global supervision.
    \item We evaluate the proposed Cross4D-JEPA on four benchmarks, MSR-Action3D~\cite{msraction3d}, DeformingThings4D~\cite{deformingthings4d}, NTU-RGB+D\,60~\cite{nturgbd}, and HOI4D~\cite{hoi4d}.
    The results show that it consistently outperforms intra-modal and global cross-modal baselines, improves label efficiency and full-label fine-tuning performance under the same budget, and matches a heavyweight pooling backbone with $13\times$ fewer parameters.
\end{itemize}

\section{Related Work}
\label{sec:related}

\noindent \textbf{Self-supervised learning on static point clouds.}
Masked autoencoding underpins much of 3D self-supervised learning, from Point-MAE~\cite{pointmae} and Point-M2AE~\cite{pointm2ae} to cross-modal variants that distill 2D knowledge into point encoders such as I2P-MAE~\cite{i2pmae}, ACT~\cite{act}, and ReCon~\cite{recon}, all of which rely on masking and decoders.
Point-JEPA~\cite{pointjepa} instead adopts the joint-embedding predictive paradigm~\cite{ijepa}, predicting target embeddings in latent space.
These methods operate on static point clouds, whereas the proposed Cross4D-JEPA adds the temporal axis and a dense per-point correspondence to a frozen 2D foundation teacher.

\vspace{3pt}
\noindent \textbf{Self-supervised learning on 4D point clouds.}
For dynamic sequences, spatio-temporal backbones such as PSTNet~\cite{pstnet}, \textsc{P4Transformer}~\cite{p4transformer}, and PPTr~\cite{pptr} support a range of pretext objectives.
These include spatio-temporal contrastive learning~\cite{strl,pointcmp,pointcpsc,fourdcontrast}, clip-order prediction~\cite{orderpred4d}, masked structure prediction~\cite{mastpre,uni4d}, and intra-modal distillation~\cite{c2p,cpr}.
The main cross-modal 4D method, CrossVideo~\cite{crossvideo}, applies contrastive losses to global clip embeddings.
The proposed Cross4D-JEPA instead introduces a dense, cross-modal, and non-contrastive objective that predicts per-patch features for a frozen 2D foundation model at each 3D point, without negatives.

\vspace{3pt}
\noindent \textbf{Cross-modal 2D-to-3D distillation.}
A separate line distills frozen 2D foundation models into 3D encoders via joint-embedding prediction, in which CrossJEPA~\cite{crossjepa} and Concerto~\cite{concerto} supervise a static point encoder using object- or scene-level embeddings from cached targets.
A complementary line transfers 2D features through an explicit 2D$\leftrightarrow$3D correspondence, using calibrated cameras and LiDAR for contrastive distillation~\cite{ppkt,slidr,seal} or per-scene feature-field optimization~\cite{openscene,dff,lerf}.
Cross-modal distillation has also been used for action recognition, transferring knowledge across modalities at the clip level~\cite{nguyen2026viewaware}.
Unlike these approaches, which require calibrated camera--LiDAR alignment or per-scene optimization, the proposed Cross4D-JEPA recovers the correspondence from geometry alone by rendering, so it needs no calibration and caches each target once; it learns a feed-forward 4D encoder rather than a static model or a per-scene field, and extends the cached-target formulation to dynamic point clouds with a dense per-point objective.

\vspace{3pt}
\noindent \textbf{Video foundation models and latent world models.}
In 2D video, masked modeling~\cite{videomae,videomaev2} and joint-embedding prediction~\cite{vjepa,vjepa2} predict masked or future latents, and recent variants act as latent world models for forecasting~\cite{thinkjepa,adljepa,adlistjepa}.
We adopt such models as teachers, namely a video model V-JEPA\,2~\cite{vjepa2} and an image model DINOv2~\cite{dinov2}, and find that the per-frame image teacher transfers densely as well as or better, while latent forecasting does not drive the gain.
By contrast, the proposed Cross4D-JEPA distills a general 3D-geometry encoder for deformable, non-driving motion, rather than the 2D-video or driving-LiDAR encoders these works produce.

\section{Method}
\label{sec:method}

\providecommand{\tok}{\mathbf{t}}                       
\providecommand{\head}{\psi}                            
\providecommand{\argmin}{\operatorname*{arg\,min}}

\begin{figure*}[t]
  \centering
  \includegraphics[width=0.95\linewidth]{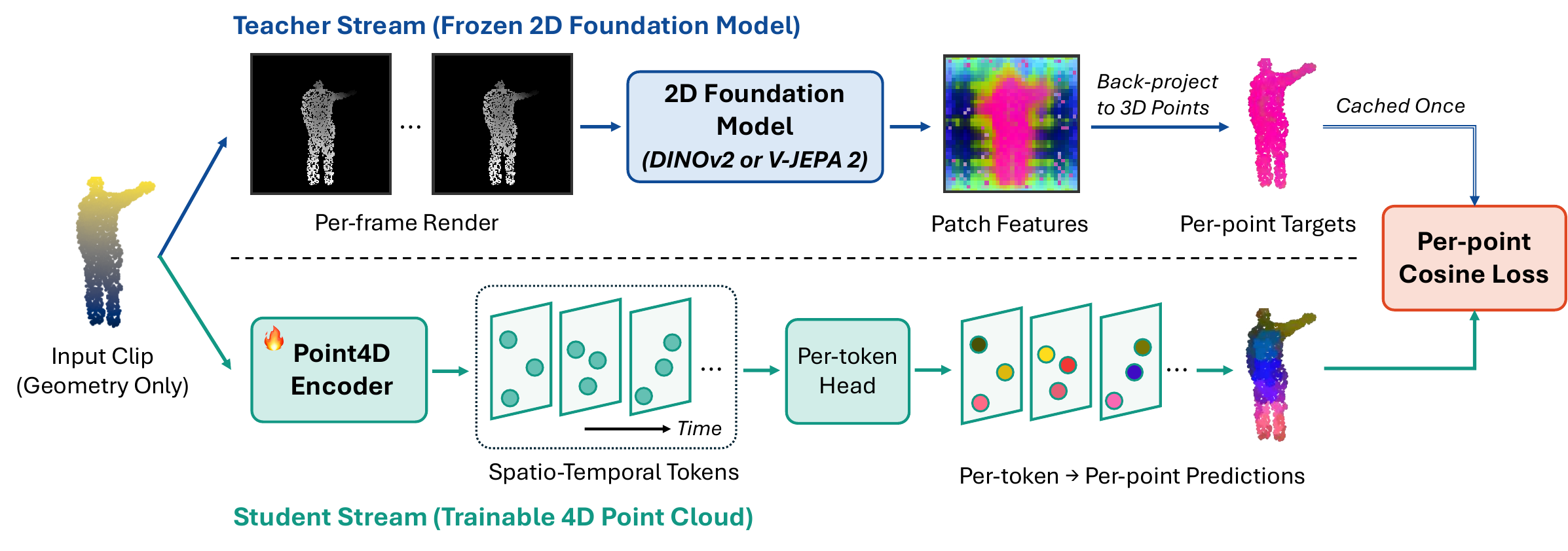}
  \vspace{-2mm}
  \caption{\textbf{Overview of Cross4D-JEPA.} A frozen 2D foundation model (top) encodes each rendered frame, and its patch features are pulled back onto the 3D points they depict, giving a per-point target cached once. A trainable 4D encoder (bottom) maps the clip to spatio-temporal tokens, and a per-token head predicts each token's target in the teacher's latent space under a cosine loss, with no masking, negatives, or decoder. The distillation is dense, per point rather than per clip.}
  \vspace{-3mm}
  \label{fig:architecture}
\end{figure*}

\noindent \textbf{Problem formulation.}
We denote a point-cloud clip as $P = \{P_f\}_{f=1}^{T}$, with per-frame point set $P_f = \{P_{f,n}\}_{n=1}^{N}\in\R^{N\times 3}$, where $T$ is the number of frames and $N$ the number of points per frame.
Self-supervised 4D representation learning seeks an encoder $\student$ that maps a clip to a set of spatio-temporal tokens $\student(P)=\{(\tok_j, c_j)\}_{j=1}^{M}$, where $\tok_j\in\R^{C}$ is a token feature and $c_j\in\R^{3}$ its spatial center, usable for downstream recognition without labels.
We learn $\student$ by distilling a frozen 2D foundation model $\teacher$, reading for each visible point a teacher feature at the pixel it projects to, forming a per-point target, and training $\student$ to predict that target in the teacher's latent space.
Unlike prior cross-modal transfer from 2D~\cite{crossvideo}, which supervises a single global embedding per clip, the supervision here is dense, with one target per point.

\vspace{3pt}
\noindent \textbf{Method overview.}
Cross4D-JEPA is a two-stream, asymmetric architecture (\cref{fig:architecture}) with three stages.
(1)~the frozen teacher $\teacher$ renders each frame and produces a per-point target $y_{f,n}$, computed once and cached;
(2)~the trainable 4D student $\student$ maps the geometry-only clip to spatio-temporal tokens $\{(\tok_j, c_j)\}_{j=1}^{M}$;
(3)~a lightweight per-token head $\head$ predicts each token's target $\hat{y}_j=\head(\tok_j)$ in the teacher's latent space, matched with a cosine loss and no masking, negatives, or decoder.
The asymmetry is deliberate, as the geometry-only student must reproduce, at every point, what the appearance-rich teacher sees at the corresponding pixel, rather than summarizing a clip into one vector.

The bridge between the 2D teacher and the 4D student is rendering, despite their different inputs.
The forward render $I_f=\mathcal{R}(P_f;\pi)$ maps the 3D geometry into the teacher's 2D input space, and the inverse point-to-pixel correspondence $\mathrm{pix}_\pi$ carries the teacher's patch features back onto the 3D points, letting an image model supervise a geometry-only encoder.
The teacher operates per frame, while the student ingests the entire clip, aggregating the per-point, per-frame targets $\{y_{f,n}\}$ into a spatio-temporal representation.
The central design choice is that this supervision is dense, with each token supervised by the teacher feature at the surface point it represents.

\subsection{Dense cross-modal correspondence}

Since the 2D teacher operates on images, we first view each frame's geometry from a single virtual viewpoint, a fixed pinhole projection $\pi=(K,[R\,|\,t])$ with intrinsics $K$ and pose $[R\,|\,t]$.
For each frame $f$, we render the points under $\pi$ into a depth-shaded image with a dependency-free, $z$-buffered splatter as:
\begin{equation}
  I_f = \mathcal{R}(P_f;\pi) \in \R^{3\times H\times W},
  \label{eq:render}
\end{equation}
placing the viewpoint at the clip centroid and backing it off to the clip's extent to keep the whole subject within the frame, regardless of dataset units.
We use a single fixed viewpoint per clip, which keeps the rendering deterministic; adding cameras raises target coverage but not probe accuracy (supplementary).
The frozen teacher maps this image to a patch-token feature map, which we compress with a fixed random projection $W\in\R^{D\times d_p}$ to bound the cache size as:
\begin{equation}
  \begin{aligned}
    \Phi_f &= \teacher(I_f) \in \R^{h_p\times w_p\times D}, \\
    \widetilde{\Phi}_f &= \Phi_f\,W \in \R^{h_p\times w_p\times d_p},
  \end{aligned}
  \label{eq:teacherfeat}
\end{equation}
where $h_p\times w_p$ is the teacher's spatial patch grid, equal to $H/14$ per side for the image teacher DINOv2~\cite{dinov2}.
For the video teacher V-JEPA\,2~\cite{vjepa2}, we run the teacher on a short rendered clip and read its spatio-temporal patch tokens per frame, which fits the same per-point construction.

To pull these features onto the geometry, let $\mathrm{pix}_\pi:\R^{3}\!\to\!\{1,\dots,h_p\}\times\{1,\dots,w_p\}$ map a 3D point to the patch it projects to under $\pi$, and let $\mathcal{V}_f\subseteq\{1,\dots,N\}$ be the points visible in frame $f$, read from an occlusion-aware $z$-buffer with deterministic tie-breaking.
The per-point target is the projected teacher feature at the corresponding patch as:
\begin{equation}
  y_{f,n} = \widetilde{\Phi}_f\big[\,\mathrm{pix}_\pi(P_{f,n})\,\big]\in\R^{d_p}, \qquad n\in\mathcal{V}_f .
  \label{eq:densetarget}
\end{equation}
Occluded points carry no target and are excluded from the loss.
Since $\teacher$ is frozen and the viewpoint is deterministic, all targets $\{y_{f,n}\}$ are precomputed once and cached in FP16, adding no teacher forward pass to the training loop.

\subsection{Student encoder and per-point prediction}

The student $\student$ is a standard 4D point backbone; we use a hierarchical PointNet++-style~\cite{qi2017pointnet,qi2017pointnetpp} spatio-temporal encoder, \textsc{Point4D}, and report the official \textsc{P4Transformer}~\cite{p4transformer} backbone where noted.
It produces the tokens $\{(\tok_j, c_j)\}_{j=1}^{M}$ of the problem formulation, each assigned to a frame $f(j)$, with $c_j$ recovered in the original (un-normalized) coordinates.
As the tokens are farthest-point-sampled (FPS)~\cite{qi2017pointnetpp} centroids that need not coincide with input points, we match each token to its nearest input point in the same frame as:
\begin{equation}
  n^{\star}(j) \;=\; \argmin_{n\in\{1,\dots,N\}}\,\bigl\lVert c_j - P_{f(j),\,n}\bigr\rVert_2 ,
  \label{eq:match}
\end{equation}
and predict its target with a lightweight linear head $\head:\R^{C}\!\to\!\R^{d_p}$, giving the prediction $\hat{y}_j=\head(\tok_j)\in\R^{d_p}$.
In our \textsc{Point4D} tokenizer, the centroids are FPS samples of the input points, so this match is near-degenerate (sub-voxel distance) and needs no distance threshold or soft assignment.

\subsection{Training objective}

Let $\mathcal{M}=\{\,j:\,n^{\star}(j)\in\mathcal{V}_{f(j)}\,\}$ be the set of tokens whose matched point is visible.
The objective is the mean negative cosine similarity between each prediction and its stop-gradient target, computed in latent space with no decoder and no negatives as:
\begin{equation}
  \mathcal{L}_{\mathrm{dense}}
  \;=\; \frac{1}{|\mathcal{M}|}\sum_{j\in\mathcal{M}}
    \Bigl(1 - \cos\!\bigl(\hat{y}_j,\;\stopgrad(y_{f(j),\,n^{\star}(j)})\bigr)\Bigr),
  \label{eq:loss}
\end{equation}
where $\cos(a,b)=a^{\top}b/(\lVert a\rVert_2\,\lVert b\rVert_2)$ and $\stopgrad$ blocks gradients to the frozen targets.
The frozen teacher provides a fixed, non-trivial target, which rules out the trivial-constant solution of EMA-target JEPAs~\cite{ijepa,vjepa}.
Because this fixed target already prevents collapse, an auxiliary anti-collapse penalty is unnecessary; we include a VICReg-style~\cite{vicreg} variance term only for completeness:
\begin{equation}
  \begin{aligned}
    \mathcal{L}_{\mathrm{var}} &= \frac{1}{d_p}\sum_{k=1}^{d_p}\mathrm{ReLU}\!\bigl(1-\sigma_k\bigr), \\
    \sigma_k &= \sqrt{\operatorname{Var}_{j\in\mathcal{M}}\!\bigl[\hat{y}_{j,k}\bigr]+\varepsilon},
  \end{aligned}
  \label{eq:var}
\end{equation}
where $\hat{y}_{j,k}$ is the $k$-th coordinate of $\hat{y}_j$ and $\varepsilon$ is a small constant for numerical stability.
We minimize $\mathcal{L}=\mathcal{L}_{\mathrm{dense}}+\lambda\,\mathcal{L}_{\mathrm{var}}$ with AdamW, linear warmup, and cosine decay; removing this term ($\lambda{=}0$) leaves results statistically unchanged (Supplementary Material), confirming the cosine target alone suffices.

\subsection{Teacher and variants}

The proposed Cross4D-JEPA is teacher-agnostic, and we consider the following choices and ablations.
\begin{itemize}
    \item \textbf{Teacher.} The default is the image foundation model DINOv2~\cite{dinov2}, whose patch features are spatially dense and semantically meaningful; we also evaluate a video teacher, V-JEPA\,2~\cite{vjepa2}, applied to short rendered clips.
    \item \textbf{Teacher input.} For \emph{all four} datasets the teacher sees the \emph{rendered} depth-shaded image (\cref{eq:render}), never the dataset's RGB; the point-to-pixel correspondence is thus exact by construction (the render's $z$-buffer index) and needs no camera calibration or RGB--depth alignment, including on the RGB-D benchmarks (NTU-RGB+D\,60~\cite{nturgbd}, HOI4D~\cite{hoi4d}).
    \item \textbf{Ablated variants.} 
    We vary two design axes. 
    Supervision granularity ranges from a single per-clip embedding to dense per-point targets, and teacher modality ranges from an image to a video model. 
    We additionally examine a latent-forecasting formulation that predicts a future view, a view-invariant latent field, a latent-dynamics rollout, and a multi-view consistency term. 
\end{itemize}

\section{Experiments}
\label{sec:experiments}

\subsection{Datasets and protocols}
\label{sec:exp:datasets}

\myparagraph{Datasets.} We evaluate the proposed method on four 4D point-cloud benchmarks spanning human, animal, and real-RGB domains, as follows:
\begin{itemize}
    \item {MSR-Action3D (MSR)}~\cite{msraction3d}: the primary benchmark for this study, $567$ depth-sensor videos over $20$ actions, under the standard cross-subject split of $270$ train and $297$ test.
    \item {DeformingThings4D-Animals {(DT4D)}}~\cite{deformingthings4d}: synthetic 4D sequences of deforming animal meshes performing non-rigid motions, $1{,}772$ animations over $38$ animal categories, under a deterministic cross-identity split of $1{,}413$ train and $359$ test.
    \item {NTU-RGB+D\,60 (NTU)}~\cite{nturgbd}: real-RGB action recognition, $56{,}880$ videos over $60$ actions, under the standard cross-subject split of $40{,}320$ train and $16{,}560$ test.
    \item {HOI4D}~\cite{hoi4d}: real RGB-D 4D semantic segmentation, $2{,}221$ annotated sequences over $\sim$$40$ categories, under the official split of $1{,}776$ train and $445$ test.
\end{itemize}

\myparagraph{Evaluation protocols.}
We report two protocols as follows:
(1)~\emph{Linear probing}: a linear head on the frozen encoder's mean-pooled features, the primary measure of representation quality;
(2)~\emph{Label-efficient fine-tuning}: end-to-end fine-tuning on a fraction $f\in\{0.1,0.25,0.5,1.0\}$ of labels, comparing an encoder initialized from Cross4D-JEPA pretraining against the identical backbone trained from scratch.
Unless noted, we report mean$\pm$std over $3$ seeds, and the gaps we highlight exceed this seed spread.

\myparagraph{Baseline methods.}
To isolate the effect of supervision \emph{granularity}, we compare three settings that share the same \textsc{Point4D} backbone, pretraining data, and probe, differing only in the distillation target, as follows:
\begin{itemize}[leftmargin=1.2em,itemsep=2pt,topsep=2pt]
    \item \emph{Random}: the same \textsc{Point4D} backbone with random initialization and no distillation, serving as the lower bound.
    \item \emph{Global}: cross-modal distillation from a single per-clip teacher embedding, the same-granularity counterpart of prior 2D-to-point transfer~\cite{crossvideo}.
    \item \emph{Dense} (ours): the per-point correspondence distillation supervises each point with its own teacher feature.
\end{itemize}

\myparagraph{Implementation details.}
The student is \textsc{Point4D}, a dependency-free PointNet++-style~\cite{qi2017pointnet,qi2017pointnetpp} spatio-temporal encoder ($3.3$M params, $T{=}24$ frames, $N{=}2048$ points), supervised by a frozen DINOv2-large teacher for $100$ epochs (AdamW~\cite{loshchilov2019decoupled}, per-point cosine objective \cref{eq:loss}) and evaluated by a linear probe on mean-pooled features.
DINOv2-large is the default teacher throughout; the teacher-modality study (\cref{sec:exp:granularity}) additionally evaluates the video models V-JEPA\,2~\cite{vjepa2} and VideoMAEv2~\cite{videomaev2}, as well as the stronger image model DINOv3~\cite{dinov3}.
For all four datasets, the teacher sees only rendered depth-shaded images (never the dataset RGB), giving calibration-free per-point targets. The full architecture, optimization, and computation, plus ablations of the random-projection dimension $d_p$, render resolution, and the VICReg term, are in the Supplementary Material.

\begin{table}[t]
    \centering
    \footnotesize
    \caption{\textbf{Effect of supervision granularity.} Comparison of the proposed dense distillation with the baselines in linear-probe accuracy on three datasets. 
    Entries are mean$\pm$std over seeds; chance denotes the random-guess accuracy.
    All distillation here uses DINOv2-large as the default teacher.}
    \vspace{-1mm}
    \label{tab:main}
    \setlength{\tabcolsep}{2.5pt}
    \renewcommand{\arraystretch}{1.2}
    \resizebox{\columnwidth}{!}
    {
    \begin{tabular}{l|c|ccc}
        \toprule
        \textbf{Dataset} & \textbf{chance} & \textbf{random} & \textbf{global} & \textbf{dense (ours)} \\
        \midrule \midrule
        MSR-Action3D            & 0.05  & 0.42\scriptsize\,$\pm$\,.02 & 0.64\scriptsize\,$\pm$\,.06            & \textbf{0.81\scriptsize\,$\pm$\,.02} \\
        DeformingThings4D & 0.03 & 0.24\scriptsize\,$\pm$\,.02 & 0.42\scriptsize\,$\pm$\,.03 & \textbf{0.49\scriptsize\,$\pm$\,.02} \\
        NTU-RGB+D\,60  & 0.02 & 0.11\scriptsize\,$\pm$\,.02 & 0.23\scriptsize\,$\pm$\,.02 & \textbf{0.40\scriptsize\,$\pm$\,.01} \\
        \bottomrule
    \end{tabular}
    }
    \vspace{-1mm}
\end{table}

\begin{table}[t]
    \centering
    \small
    \caption{\textbf{Granularity versus teacher modality} in linear-probe accuracy on the MSR-Action3D dataset with the \textsc{Point4D} backbone, given as mean$\pm$std.
    }
    \vspace{-1mm}
    \label{tab:modality}
    \setlength{\tabcolsep}{3pt}
    \renewcommand{\arraystretch}{1.2}
    \begin{tabular}{l|c|cc}
        \toprule
        \textbf{Teacher} & \textbf{modality} & \textbf{global} & \textbf{dense (ours)} \\
        \midrule \midrule
        DINOv2        & image & 0.64\scriptsize\,$\pm$\,.06 & \textbf{0.81\scriptsize\,$\pm$\,.02} \\
        DINOv3        & image & 0.64\scriptsize\,$\pm$\,.02  & \textbf{0.87\scriptsize\,$\pm$\,.03} \\
        V-JEPA\,2     & video & 0.61\scriptsize\,$\pm$\,.08 & \textbf{0.88\scriptsize\,$\pm$\,.01} \\
        VideoMAEv2 & video & 0.63\scriptsize\,$\pm$\,.07 & \textbf{0.81\scriptsize\,$\pm$\,.01} \\
        \bottomrule
    \end{tabular}
    \vspace{-3mm}
\end{table}

\subsection{Dense distillation is the lever}
\label{sec:exp:main}
\begin{figure}[t]
  \centering
  \includegraphics[width=0.95\linewidth]{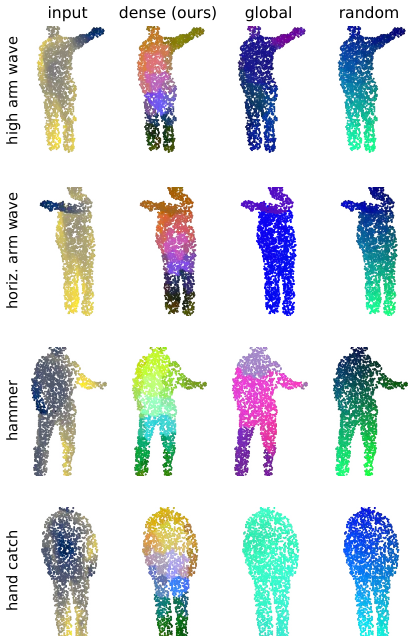}
  \caption{\textbf{Dense distillation paints coherent body-part semantics onto 3D points.} 
  Qualitative counterpart of \cref{tab:main} on four MSR-Action3D actions (rows). 
  Each point is colored by the PCA-to-RGB of its nearest token from the \textbf{dense (ours)}, \emph{global}, and \emph{random} encoders, shown next to the input. 
  Only the dense encoder gives the same body part a consistent color across poses without point-level supervision, whereas the global encoder collapses to a near-uniform per-clip color, and random is unstructured. 
  }
  \vspace{-3mm}
  \label{fig:dense_features}
\end{figure}

Table~\ref{tab:main} reports the linear-probe accuracy of the random, global, and dense encoders, given as mean$\pm$std.
On the MSR-Action3D dataset, the proposed dense distillation achieves $0.81$, surpassing the global baseline by $0.17$ and the random encoder by $0.39$ with the same probe.
This $0.17$ gain exceeds the inter-seed spread --- the dense and global error bars do not overlap --- confirming that distilling \emph{where} each 2D feature belongs, rather than one vector per clip, makes the representation linearly separable.
The trend is consistent across all three datasets: dense improves over global by $0.17$, $0.07$, and $0.17$ on MSR-Action3D, DeformingThings4D, and NTU-RGB+D, with non-overlapping error bars.
The lower absolute accuracies on the latter two reflect their larger label spaces of $38$ and $60$ classes compared to MSR-Action3D's $20$. Yet, all results remain $16$ to $20\times$ the chance, and the dense-over-global ordering holds across all three datasets.
\Cref{fig:dense_features} provides the qualitative counterpart of this ordering.

\subsubsection{Effect of teacher modality}
\label{sec:exp:granularity}
Table~\ref{tab:modality} varies the teacher to test whether the gain depends on its modality rather than on the dense correspondence.
The dense objective is teacher-agnostic: under global supervision, every teacher sits in the $0.61$ to $0.64$ range regardless of modality, with the image and video teachers tied within their error bars, whereas dense supervision lifts each one into the $0.81$ to $0.88$ range, so a particular teacher is not what drives the transfer.
Modality acts only as a secondary lever once supervision is dense, where the video teacher V-JEPA\,2 reaches $0.88$ and the stronger image teacher DINOv3 reaches $0.87$, both above DINOv2 and VideoMAEv2 at $0.81$; this $0.07$ spread among teachers is far smaller than the global-to-dense step of at least $0.17$, so the secondary gain tracks the teacher's dense-feature quality rather than its modality, and we adopt DINOv2 as the default for its clean per-patch correspondence.
We similarly rule out the forecasting pretext tasks common in prior 4D self-supervised learning, such as future-view and latent-dynamics prediction: the variants we explored never beat dense distillation's $0.81$, topping out at $0.33$ on a lightweight backbone and remaining statistically indistinguishable from one another on \textsc{P4Transformer}, confirming that the per-point correspondence is the operative mechanism.

\subsubsection{Effect of backbone resolution}
\label{sec:exp:backbone}
\Cref{tab:granxback} shows that the granularity lever interacts with the encoder's token resolution.
The \textsc{Point4D} backbone, with $3.3$M parameters, keeps tokens close to individual surface points and can absorb per-point targets, so dense supervision lifts it to $0.81$ while global supervision leaves it at $0.64$.
The official \textsc{P4Transformer}, with $42$M parameters, instead pools each clip toward a single classification token, which inverts the ordering: global supervision reaches $0.80$, but under dense supervision the probe falls to $\sim$$0.60$, consistent with its pooled tokenization discarding the per-point targets, as detailed in the supplementary material.
Each backbone is therefore best at its \emph{own} token granularity, and the two bold cells are directly comparable.
The practical payoff is that dense distillation lets a $13\times$ smaller backbone match the heavyweight pooling backbone's accuracy while retaining the per-point features shown in \cref{fig:dense_features}.
The dense $\gg$ global ordering thus reflects a per-point-preserving backbone, not a claim that global supervision is generally weak.

\begin{table}[t]
    \centering
    \small
    \caption{\textbf{Granularity $\times$ backbone.} Linear-probe accuracy on the MSR-Action3D dataset with the DINOv2 teacher. The bold cells are comparable, but only dense supervision with \textsc{Point4D} yields per-point features, at $13\times$ fewer parameters.}
    \label{tab:granxback}
    \setlength{\tabcolsep}{5pt}
    \renewcommand{\arraystretch}{1.2}
    \vspace{-1mm}
    \begin{tabular}{l|cc}
        \toprule
        \textbf{Backbone} & \textbf{global} & \textbf{dense (ours)} \\
        \midrule \midrule
        \textsc{Point4D} ($3.3$M)        & 0.64 & \hspace{5pt} \textbf{0.81} \\
        \textsc{P4Transformer} ($42$M)   & \textbf{0.80} & $\sim$0.60 \\
        \bottomrule
    \end{tabular}
    \vspace{-1mm}
\end{table}

\begin{table}[t]
    \centering
    \small
    \caption{\textbf{Label-efficient fine-tuning} on the MSR-Action3D dataset with the \textsc{Point4D} backbone, given as mean$\pm$std.}
    \label{tab:labeleff}
    \setlength{\tabcolsep}{5pt}
    \renewcommand{\arraystretch}{1.2}
    \vspace{-1mm}
    \begin{tabular}{l|cccc}
        \toprule
        \textbf{Labels} & \textbf{10\%} & \textbf{25\%} & \textbf{50\%} & \textbf{100\%} \\
        \midrule \midrule
        From scratch        & 0.29{\scriptsize$\pm$.11} & 0.63{\scriptsize$\pm$.02} & 0.76{\scriptsize$\pm$.01} & 0.82{\scriptsize$\pm$.01} \\
        Dense (ours)   & \textbf{0.46}{\scriptsize$\pm$.04} & \textbf{0.73}{\scriptsize$\pm$.02} & \textbf{0.88}{\scriptsize$\pm$.02} & \textbf{0.92}{\scriptsize$\pm$.01} \\
        \midrule
        $\Delta$            & +0.17 & +0.10 & +0.12 & +0.10 \\
        \bottomrule
    \end{tabular}
    \vspace{-2mm}
\end{table}

\subsection{Label-efficient fine-tuning}
\label{sec:exp:labeleff}
Table~\ref{tab:labeleff} reports fine-tuning from the dense encoder against training from scratch at four label fractions on the MSR-Action3D dataset.
The dense initialization wins across all fractions, and the gain is largest when labels are scarce, at $+0.17$ with $10\%$, where the pretrained part structure serves as a proxy for the missing supervision.
Unlike pretraining gains that usually fade once labels are plentiful, the advantage persists at full supervision, still adding $+0.10$ at $100\%$ and reaching $0.92$ against $0.82$ with non-overlapping error bars, which shows that the dense correspondence encodes structure that label supervision alone does not recover and thereby raises the fully-supervised ceiling.
The $+0.17$ at $10\%$ also exceeds the $+0.14$ from global distillation, so the benefit again traces to per-point granularity.

\subsection{Generality and robustness}
\label{sec:exp:generality}

\begin{table}[t]
    \centering
    \small
    \caption{\textbf{Cross-domain transfer.} Linear-probe accuracy of a frozen encoder pretrained on the source row and probed on the target column with no target-domain pretraining. The diagonal cells are in-domain, where the source and target match. The bottom row, \emph{random}, is a randomly initialized \textsc{Point4D} encoder probed on each target and serves as the lower baseline that every cross-domain cell exceeds.}
    \label{tab:transfer}
    \setlength{\tabcolsep}{4pt}
    \renewcommand{\arraystretch}{1.2}
    \vspace{-1mm}
    {
    \begin{tabular}{l|ccc}
        \toprule
        \textbf{Source $\downarrow$ / Target $\to$} & \textbf{MSR} & \textbf{DT4D} & \textbf{NTU} \\
        \midrule \midrule
        MSR-Action3D       & 0.81 & 0.44 & 0.46 \\
        DeformingThings4D  & 0.67 & 0.49 & 0.32 \\
        NTU-RGB+D\,60      & 0.68 & 0.44 & 0.40 \\
        \midrule
        random             & 0.59 & 0.27 & 0.25 \\
        \bottomrule
    \end{tabular}
    }
    \vspace{-2mm}
\end{table}

\begin{figure}[t]
    \centering
    \includegraphics[width=0.95\linewidth]{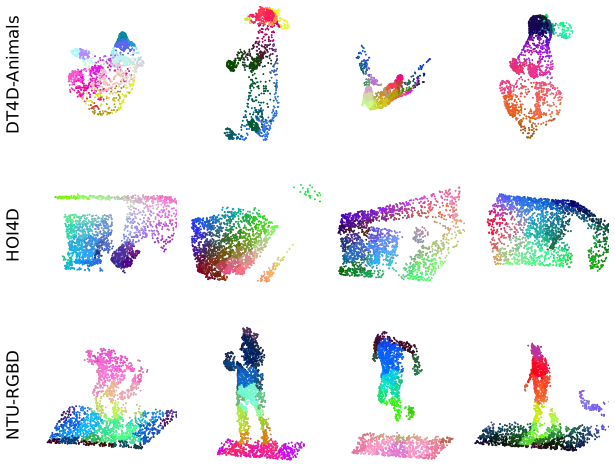}
    \caption{\textbf{Dense features generalize across domains.} 
    Per-point features of the frozen dense encoder {pretrained on MSR-Action3D and applied to} DeformingThings4D-Animals, HOI4D, and NTU-RGB+D, with each point colored by its per-clip PCA-to-RGB.
    Part-coherent regions emerge across the animal, hand-object, and real-RGB human domains, not only on the MSR-Action3D source.
    }
    \label{fig:crossdataset}
    \vspace{-2mm}
  \end{figure}

\myparagraph{Cross-domain generality.} Table~\ref{tab:transfer} probes a frozen dense encoder pretrained on one dataset on \emph{another}, with no target-domain pretraining, and it beats a random encoder on every source-to-target pair across the human, animal, and real-RGB domains. The gains are largest for the harder targets, for example, $0.46$ against $0.25$ for MSR-to-NTU and $0.44$ against $0.27$ for MSR-to-DT4D, and even on MSR, where a random encoder already reaches $\sim$$0.59$, the transferred encoders still improve to $\sim$$0.67$. \Cref{fig:crossdataset} shows the same generality qualitatively. Dense distillation thus learns transferable structure rather than dataset-specific shortcuts.

\begin{table}[t]
    \centering
    \footnotesize
    \caption{\textbf{HOI4D 4D semantic segmentation.} {Per-point linear probe on a frozen encoder reporting mIoU/mAcc/OA in \% on a subsampled $24$-frame/$2048$-point split with $\sim$$40$ classes.}}
    \vspace{-1mm}
    \label{tab:hoi4dseg}
    \setlength{\tabcolsep}{6pt}
    \renewcommand{\arraystretch}{1.2}
    \begin{tabular}{l|ccc}
        \toprule
        \textbf{Encoder} & \textbf{mIoU} & \textbf{mAcc} & \textbf{OA} \\
        \midrule \midrule
        random init                  & \hspace{2pt} 4.6  & \hspace{2pt} 7.3  & 48.0 \\
        \midrule
        \multicolumn{4}{l}{\textit{Zero-shot transfer (pretrained on MSR-Action3D)}}\\
        \quad global                 & \hspace{2pt} 4.4  & \hspace{2pt} 7.2  & 43.7 \\
        \quad \textbf{dense (ours)}  & \textbf{16.2} & \textbf{21.4} & \textbf{59.4} \\
        \midrule
        \multicolumn{4}{l}{\textit{In-domain (pretrained on HOI4D)}}\\
        \quad global                 & \hspace{2pt} 7.2  & 10.8 & 48.5 \\
        \quad \textbf{dense (ours)}  & \textbf{23.6} & \textbf{32.5} & \textbf{65.3} \\
        \bottomrule
    \end{tabular}
\end{table}

\begin{figure}[t]
\centering
\includegraphics[width=0.95\linewidth]{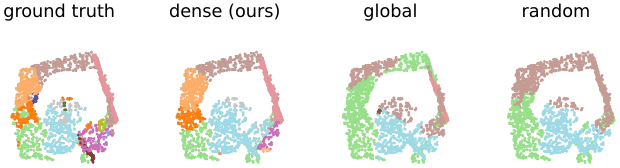}
\caption{
\textbf{Dense distillation recovers per-point HOI4D segmentation.} 
One HOI4D validation clip, with each point colored by its ground-truth or in-domain-predicted semantic class under a shared palette, where matching colors indicate correct predictions, and gray is the ignored background.}
\label{fig:semseg_qual}
\vspace{-3mm}
\end{figure}

\myparagraph{Zero-shot 4D semantic segmentation.} 
The most direct test of a dense objective is per-point labeling itself.
\Cref{tab:hoi4dseg} reports HOI4D 4D semantic segmentation with a frozen encoder and a linear per-point head.
Transferred zero-shot from MSR-Action3D, the dense encoder reaches $16.2$ mIoU, $3.5\times$ random init, and $3.7\times$ the global encoder, which itself sits at the random floor, with part-coherent regions emerging on the unseen HOI4D geometry under no target-domain labels (\cref{fig:crossdataset}).
Pretrained in-domain, dense rises to $23.6$ against $7.2$ for global, and the dense encoder's per-point predictions closely track the ground truth on a held-out clip, whereas global and random collapse to a few classes (\cref{fig:semseg_qual}).
Per-point supervision is thus what enables the spatio-temporal tokens to carry transferable per-point semantics, whereas clip pooling discards them.
The effect also persists under full supervision, where initializing an end-to-end segmenter from the dense encoder raises the fully-supervised ceiling from $39.0$ to $40.9$ mIoU ($+1.9$, $3$ seeds;  in the Supplementary Material).

\begin{figure}[t]
  \centering
  \includegraphics[width=0.95\linewidth]{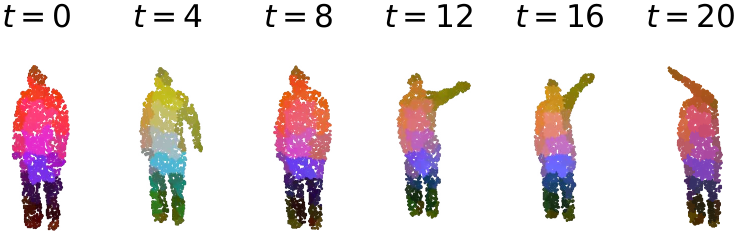}
  \caption{\textbf{Temporal robustness of the dense features.} One MSR-Action3D motion across frames $t{=}0$ to $20$, each point colored by a single shared PCA-to-RGB of the dense encoder's per-point features. 
  Body-part regions keep a consistent color through the motion. 
  The claim is region-level color stability, not a frozen segmentation, as MSR-Action3D resamples points per frame.
  }
  \label{fig:temporal}
  \vspace{-1mm}
\end{figure}

\myparagraph{Temporal robustness.} 
On an in-domain clip, the dense features are stable over time, keeping each body part a consistent color across the frames of a motion, as \cref{fig:temporal} shows. 
The per-point representation thus tracks parts through the deformation rather than drifting frame to frame.

\myparagraph{Test-time robustness.}
Under test-time input corruption with no retraining, the dense features are far more resilient than a random encoder to temporal subsampling, still reaching $0.35$ at $6$ of $24$ frames against $0.04$, and degrade gracefully under point dropout and coordinate jitter. 
The full sweeps are in the Supplementary Material.

\begin{table}[t]
    \centering
    \footnotesize
    \caption{\textbf{Comparison with published 4D self-supervised learning.} MSR-Action3D linear-probe accuracy. \emph{Top}: published numbers on their native \textsc{P4Transformer} backbone. 
    \emph{Bottom}: each pretext reimplemented on the \textsc{Point4D} backbone with identical data and probe, at its native training budget. 
    }
    \vspace{-1mm}
    \label{tab:sota}
    \setlength{\tabcolsep}{4pt}
    \renewcommand{\arraystretch}{1.2}
    \begin{tabular}{l|l|c}
        \toprule
        \textbf{Method} & \textbf{Backbone} & \textbf{LP acc.} \\
        \midrule \midrule
        \multicolumn{3}{l}{\textit{Published (native backbone; not matched)}}\\
        \quad MaST-Pre~\cite{mastpre}        & \textsc{P4Transformer} & 0.64 \\
        \quad Uni4D~\cite{uni4d}             & \textsc{P4Transformer} & \textbf{0.85} \\
        \midrule
        \multicolumn{3}{l}{\textit{Matched on the \textsc{Point4D} backbone (same data + probe)}}\\
        \quad random init                    & \textsc{Point4D} & 0.42 \\
        \quad Uni4D pretext                  & \textsc{Point4D} & 0.25 \\
        \quad MaST-Pre pretext               & \textsc{Point4D} & 0.33 \\
        \quad i-JEPA (intra-modal latent)    & \textsc{Point4D} & 0.55 \\
        \quad \textbf{Cross4D-JEPA (dense)}  & \textsc{Point4D} & \textbf{0.81} \\
        \bottomrule
    \end{tabular}
    \vspace{-2mm}
\end{table}

\subsection{{Comparison with 4D self-supervised methods}}
\label{sec:exp:context}

\Cref{tab:sota} places the proposed method in the context of published 4D Self-Supervised Learning (SSL) methods, which pretrain a 4D encoder on unlabeled point sequences. 
These methods differ in their backbones, corpora, and protocols so that a raw head-to-head comparison would confound the granularity effect studied here; we therefore report two complementary views.
First, against published numbers on their native \textsc{P4Transformer} backbone, dense distillation on the lighter \textsc{Point4D} already exceeds MaST-Pre~\cite{mastpre} and approaches Uni4D~\cite{uni4d} despite a weaker encoder, and its fine-tuned ceiling ($0.92$, \cref{tab:labeleff}) is competitive with from-scratch supervision, although no fully-supervised state-of-the-art claim is made. 
Second, under a matched protocol that shares the \textsc{Point4D} backbone, pretraining data, and probe, dense distillation reaches $0.81$ and outperforms every intra-modal pretext by a wide margin, namely an i-JEPA-style masked-latent objective ($0.55$) and reimplementations of MaST-Pre ($0.33$) and Uni4D ($0.25$).
These pretexts are native to \textsc{P4Transformer}; reimplemented on \textsc{Point4D} at their published budget and without tuning their unpublished loss weights, an untuned intra-modal pretext can leave features no more linearly separable than random initialization, so their absolute values warrant caution.
The consistent gap, rather than the exact baseline values, is the robust conclusion: on a per-point backbone, it is the 2D-foundation teacher, not the self-supervised framework, that drives the transfer.

\section{Discussion}
\label{sec:discussion}
\vspace{-2mm}

\myparagraph{Why dense supervision transfers and global does not.}
We attribute the gap between dense and global supervision to where the teacher's structure is forced to land.
A single per-clip embedding routes all of the teacher's spatial structure through a single vector, which a student can match with a coarse, pose-agnostic summary; thus, on a per-point backbone, we see global distillation plateau near its global-supervision ceiling ($\approx0.64$).
Dense supervision instead requires the student to reproduce the teacher's feature at \emph{each surface point}, placing localized appearance and part semantics onto the geometry rather than averaging them away; we find this per-point structure is what makes actions linearly separable ($0.81$ vs.\ $0.64$ at identical teacher, backbone, and budget).
We read the same effect as the reason \emph{granularity}, not \emph{modality}, is the dominant lever: supervision granularity sets the first-order effect ($0.64\!\to\!0.81$), whereas the teacher's dense-feature quality, not its modality, adds only a secondary gain once dense correspondence is in place --- a stronger \emph{image} teacher matches the video teacher.

\myparagraph{Dense distillation needs a backbone that preserves per-point resolution.}
We observe that the same objective on \textsc{P4Transformer} reaches only $\approx0.60$ (\cref{sec:exp:backbone}).
We read this as consistent with the backbone's pooled tokenization rather than an optimization failure: bounding the feature scale fixes training but not the probe, while the pooling toward a single classification token leaves little per-point structure for the dense objective.
That same pooling conversely makes \textsc{P4Transformer} \emph{strong} under \emph{global} supervision ($0.80$): supervision granularity must match the granularity the backbone preserves, so we regard ``dense\,$\gg$\,global'' as specific to a per-point backbone.
The payoff is concrete, as this pairing matches the heavyweight pooling configuration's accuracy with $13{\times}$ fewer parameters while retaining the per-point features that a pooled clip token discards.

\myparagraph{Relation to distilled feature fields.}
We view lifting a 2D foundation model's features into 3D via correspondence as a 4D analog of distilled feature fields, yet more scalable: the field is obtained in one feed-forward pass of a geometry encoder rather than via per-scene optimization, and we cache the targets once, adding no teacher forward pass to training.

\section{Conclusion}
\label{sec:conclusion}

We proposed Cross4D-JEPA, a cross-modal method that distills a frozen 2D foundation model into a 4D point encoder \emph{densely}, by rendering the geometry, recovering an occlusion-aware 2D$\leftrightarrow$3D correspondence, and supervising each encoder token with the teacher patch feature of the surface point it represents, in latent space, with no masking, contrastive negatives, or decoder.
Across four 4D benchmarks, dense distillation consistently outperforms intra-modal and global cross-modal baselines, transfers across domains, improves label efficiency, and matches a heavyweight pooling backbone at a fraction of its size.
The central insight is that correspondence granularity, not teacher modality, drives the transfer: supervising where each 2D feature belongs, per point rather than per clip, places localized part semantics onto the geometry and makes the representation broadly useful.

\myparagraph{Limitations.}
We center the multi-seed study on MSR-Action3D with a dependency-free \textsc{Point4D} backbone; the official \textsc{P4Transformer} underperforms the dense objective, since its aggressive pooling discards the per-point structure dense distillation needs (supplementary material).
On the large NTU-RGB+D 60, we pretrain on a clip subset to fit the in-memory target cache.
Dense distillation presupposes a teacher with spatially dense features and a render that places the subject in the frame; for depth-only data, we approximate real appearance with the rendered views.

\myparagraph{Future work.}
The per-point targets inherit the teacher's \emph{semantic} feature space, visible as coherent within-frame regions (\cref{fig:dense_features}). Because the pipeline is teacher-agnostic and its render-based correspondence needs no calibration, substituting a \emph{dense} language-aligned teacher (e.g., a dense CLIP variant) is a direct drop-in that reuses the same cached machinery to enable open-vocabulary 3D queries over point sequences.
We note that these features capture within-frame semantics but are not yet \emph{pose-invariant}: across large deformations, geometry alone remains a strong baseline for cross-frame correspondence, so we leave open the question of learning pose-invariant dense 3D descriptors, for example, with correspondence-aware objectives or a finer backbone.
Beyond rendered or depth views, we see recovering large-scale dynamic point clouds from real video with reconstruction pipelines such as Stereo4D~\cite{stereo4d} and DePT3R~\cite{dept3r} as a promising route to scale dense cross-modal pretraining beyond curated meshes.

\section*{Acknowledgements}
This work used the Delta system at the National Center for Supercomputing Applications (NCSA) through allocation CIS260951 from the Advanced Cyberinfrastructure Coordination Ecosystem: Services \& Support (ACCESS) program, which is supported by U.S. National Science Foundation grants \#2138259, \#2138286, \#2138307, \#2137603, and \#2138296.


\appendix

\section{Additional implementation and robustness detail}
\label{supp:detail}

This section collects the dataset statistics, training defaults, and the design-choice and test-time robustness analyses referenced from the main paper.

\myparagraph{Datasets.} \textbf{MSR-Action3D}: $567$ depth-sensor videos over $20$ actions, standard cross-subject split ($270$ train / $297$ test). \textbf{DeformingThings4D-Animals}: synthetic 4D sequences of deforming animal meshes, $1{,}772$ animations over $38$ animal categories, deterministic cross-identity split ($1{,}413$ / $359$). \textbf{NTU-RGB+D\,60}: real-RGB action recognition, $56{,}880$ videos over $60$ actions, standard cross-subject split ($40{,}320$ / $16{,}560$). \textbf{HOI4D}: real RGB-D 4D semantic segmentation, $2{,}221$ annotated sequences over $\sim$$40$ categories, official split ($1{,}776$ / $445$).

\myparagraph{Architecture and training defaults.} Table~\ref{tab:trainingdefaults} lists the per-dataset clip settings. The student \textsc{Point4D} ($3.3$M params) is a PointNet++-style encoder with two set-abstraction levels and a depth-$4$, $4$-head spatio-temporal transformer at feature dimension $256$. The frozen DINOv2-large teacher's patch tokens are randomly projected to $d_p{=}384$ and cached in fp16; rendered images are $126{\times}126$ ($9{\times}9$ DINOv2 patches), one fixed teacher camera per clip. We pretrain with AdamW (learning rate $5{\times}10^{-4}$, weight decay $10^{-4}$, $5\%$ linear warmup then cosine decay), optimizing the per-point cosine loss with a small VICReg term. The linear probe runs $300$ epochs on mean-pooled features without standardization; light point augmentation (scaling, $z$-rotation, jitter) is applied only during downstream fine-tuning. All runs use a single NVIDIA A100 or H200 GPU, and the encoder is identical across pretraining and downstream tasks, thereby isolating the self-supervised gain.

\begin{table}[t]
    \centering
    \small
    \caption{Training defaults.}
    \label{tab:trainingdefaults}
    \setlength{\tabcolsep}{2pt}
    \renewcommand{\arraystretch}{1.2}
    \begin{tabular}{l|c|c|c|c}
        \toprule
        \textbf{Dataset} & \textbf{Frames} $\bm{T}$ & \textbf{Points} $\bm{N}$ & \textbf{Stride} & \textbf{Pretrain ep.} \\
        \midrule \midrule
        MSR-Action3D      & 24 & 2048 & 1 & $\sim$100 \\
        DeformingThings4D & 16 & 2048 & 1 & $\sim$60  \\
        NTU-RGBD 60       & 24 & 2048 & 2 & $\sim$100 \\
        HOI4D             & 24 & 2048 & 1 & $\sim$60  \\
        \bottomrule
    \end{tabular}
\end{table}

\myparagraph{Full-split NTU pretraining.} The main-table NTU-RGB+D result pretrains on a $1{,}000$-clip subset to bound the in-memory target cache. Using the token-granularity cache, we also pretrain on the \emph{full} $40{,}320$-clip Cross-Subject training split: the dense encoder's linear probe reaches $0.43$ (Cross-Subject) and $0.40$ (Cross-View) at a single seed, at or above the subset's $0.40$. The subset is therefore a conservative estimate, and the dense-over-global advantage is not an artifact of the reduced corpus.

\myparagraph{Efficiency and computation.} Table~\ref{tab:efficiency} reports parameters and FLOPs per clip; the student is small, and the frozen teacher is cached (no teacher forward pass in the training loop). Pretraining is cache-amortized: teacher patch features and the per-point correspondence are rendered and cached once per dataset, then read as FP16. Because targets are stored at the student's token granularity (not per raw point), the cache is tiny --- $7$\,MB for MSR-Action3D, $25$--$35$\,MB for HOI4D and DeformingThings4D --- and end-to-end pretraining (target precompute $+$ $100$ epochs, all clips) completes in $\sim\!35$ minutes on a single GPU.

\begin{table}[t]
    \centering
    \small
    \caption{\textbf{Efficiency.} Parameters and FLOPs for one clip ($24{\times}2048{\times}3$). \textsc{P4Transformer} is shown for context.}
    \label{tab:efficiency}
    \setlength{\tabcolsep}{6pt}
    \renewcommand{\arraystretch}{1.2}
    \begin{tabular}{l|cc}
        \toprule
        \textbf{Component} & \textbf{Params (M)} & \textbf{FLOPs (G)} \\
        \midrule \midrule
        \textsc{Point4D} encoder (ours)   & 3.28 & 11.5 \\
        \quad $+$ dense per-token head     & 0.10 & 0.30 \\
        \midrule
        \textsc{P4Transformer} (context)  & 42.0 & 77.1 \\
        \bottomrule
    \end{tabular}
\end{table}

\begin{figure}[t]
  \centering
  \includegraphics[width=\linewidth]{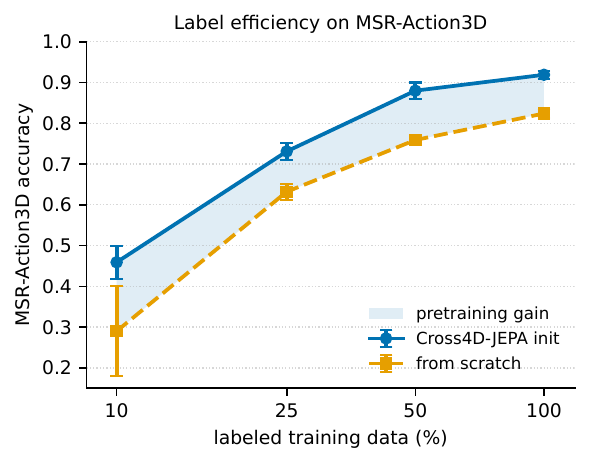}
  \caption{\textbf{Label efficiency on MSR-Action3D.} Action-recognition accuracy versus the fraction of labeled training clips, for a from-scratch encoder (orange) and the same architecture initialized from Cross4D-JEPA pretraining (blue); error bars are over seeds and the shaded band is the pretraining gain. The dense-distilled initialization helps most when labels are scarce ($+17$ points at $10\%$) and remains ahead at full supervision ($91.9\%$ vs.\ $82.4\%$).}
  \label{fig:labeleff}
\end{figure}

\myparagraph{Label efficiency (plot).} Figure~\ref{fig:labeleff} plots the MSR-Action3D label-efficiency result reported in the main paper: the dense-distilled initialization helps most when labels are scarce and remains ahead at full supervision.

\myparagraph{Sensitivity to design choices.} The dense result is stable across the random-projection dimension $d_p$ and the render resolution (Table~\ref{tab:sensitivity}): every setting lands in $0.81$--$0.85$, well above the $0.64$ global baseline, with higher render resolution helping slightly. The method is not hyperparameter-fragile. Replacing the random projection with a learned PCA projection of the same dimensionality leaves the probe essentially unchanged ($0.82$ vs.\ $0.81$), consistent with the Johnson--Lindenstrauss property that a random projection approximately preserves the cosine geometry used by the loss; we therefore keep the cheaper, learning-free random projection.

\begin{table}[t]
    \centering
    \small
    \caption{\textbf{Sensitivity} of MSR-Action3D dense linear-probe accuracy to the projection type, the random-projection dimension $d_p$, and render image size (one seed; default random $d_p{=}384$, img $126$). All settings remain in the $0.81$--$0.85$ range, far above the $0.64$ global baseline.}
    \label{tab:sensitivity}
    \setlength{\tabcolsep}{6pt}
    \renewcommand{\arraystretch}{1.2}
    \begin{tabular}{l|c}
        \toprule
        \textbf{Setting} & \textbf{LP acc.} \\
        \midrule \midrule
        default ($d_p{=}384$, img $126$) & 0.81 \\
        $d_p{=}768$                      & 0.807 \\
        PCA projection ($d_p{=}384$)     & 0.82 \\
        render img $98$                  & 0.807 \\
        render img $168$                 & 0.847 \\
        \bottomrule
    \end{tabular}
\end{table}

\myparagraph{Feature standardization helps the random baseline, not the method.} Our linear probe uses \emph{no} feature standardization. Table~\ref{tab:standardize} reports both settings: z-scoring the frozen features before the probe lifts the \emph{random-init} baseline from $0.40$ to $0.59$ (uninformative features become more separable once normalized), but slightly \emph{lowers} the dense encoder ($0.80\!\to\!0.79$). The no-standardization choice is therefore \emph{conservative} --- it does not help our method, only deflates the baseline it is measured against --- and dense\,$\gg$\, random holds under both ($+0.39$ off, $+0.19$ on).

\begin{table}[t]
    \centering
    \small
    \caption{\textbf{Probe feature standardization on/off} (MSR-Action3D linear-probe acc., single seed). Standardization inflates the \emph{random} baseline by $+0.19$ but slightly hurts the \emph{dense} encoder; we report the conservative std-off setting throughout. Dense\,$\gg$\, random under both.}
    \label{tab:standardize}
    \setlength{\tabcolsep}{8pt}
    \renewcommand{\arraystretch}{1.2}
    \begin{tabular}{l|cc}
        \toprule
        \textbf{Encoder} & \textbf{std off (default)} & \textbf{std on} \\
        \midrule \midrule
        random init   & 0.40 & 0.59 \\
        dense (ours)  & 0.80 & 0.79 \\
        \bottomrule
    \end{tabular}
\end{table}

\myparagraph{Cross-domain transfer (full matrix).} A \emph{frozen} dense encoder pretrained on one dataset and linear-probed on \emph{another} --- with no target-domain pretraining --- beats a random encoder on every source$\to$target pair (Table~\ref{tab:transferfull}). Gains are largest for the harder targets (MSR\,$\to$\,NTU $0.463$ vs.\ $0.254$, MSR\,$\to$\,DT4D $0.435$ vs.\ $0.265$, NTU\,$\to$\,DT4D $0.440$); on MSR, where random init already reaches $\sim$0.59, cross-domain encoders still beat it ($\approx0.67$--$0.68$). Dense distillation thus learns transferable structure rather than dataset-specific shortcuts.

\begin{table}[t]
    \centering
    \small
    \caption{\textbf{Cross-domain transfer matrix} (linear-probe acc.): a \emph{frozen} encoder pretrained on the source (row) is probed on the target (column) with no target-domain pretraining. \emph{Diagonal} (italic) is in-domain. Every cross-domain cell outperforms the target's random baseline across the human (MSR), animal (DT4D), and real-RGB (NTU) domains.}
    \label{tab:transferfull}
    \setlength{\tabcolsep}{6pt}
    \renewcommand{\arraystretch}{1.2}
    \begin{tabular}{l|ccc}
        \toprule
        \textbf{Source $\downarrow$ \,/\, Target $\to$} & \textbf{MSR} & \textbf{DT4D} & \textbf{NTU} \\
        \midrule \midrule
        MSR        & \textit{0.826} & 0.435 & 0.463 \\
        DT4D       & 0.666 & \textit{0.490} & 0.324 \\
        NTU        & 0.680 & 0.440 & \textit{0.400} \\
        \midrule
        random     & 0.593 & 0.265 & 0.254 \\
        \bottomrule
    \end{tabular}
\end{table}

\myparagraph{Test-time input corruption.} With no retraining, we probe the frozen dense vs.\ random encoders under test-time input perturbations (Table~\ref{tab:robust}). Dense pretraining is far more robust to \emph{temporal subsampling} --- at $6$ of $24$ frames it still reaches $0.35$ while a random encoder collapses to $0.04$ --- and degrades gracefully under moderate point dropout and coordinate jitter. The one regime where dense loses its edge is \emph{severe} corruption ($25\%$ of points, $\sigma{=}0.05$): because the dense features genuinely encode geometry, destroying the geometry hurts them more than the already-uninformative random features. The advantage is thus a property of the learned structure --- present exactly where structure survives.

\begin{table}[t]
    \centering
    \small
    \caption{\textbf{Test-time robustness} (MSR-Action3D linear-probe acc.; the probe head is trained on clean features and evaluated on \emph{perturbed} test inputs, no retraining). Single representative run per encoder --- the clean (``none'') values lie within the per-seed range behind the main result, and the point is the \emph{degradation trend}, not the absolute clean value. Dense degrades gracefully and is far more robust to temporal subsampling than random; under \emph{severe} corruption (rightmost of each block), the geometry-dependent dense features lose their edge.}
    \label{tab:robust}
    \setlength{\tabcolsep}{4pt}
    \renewcommand{\arraystretch}{1.15}
    \begin{tabular}{l|cccc}
        \toprule
         & \textbf{none} & \textbf{mild} & \textbf{mod.} & \textbf{severe} \\
        \midrule \midrule
        \multicolumn{5}{l}{\emph{temporal subsample} ($24/18/12/6$ frames)} \\
        \quad dense  & 0.85 & 0.81 & 0.69 & 0.35 \\
        \quad random & 0.40 & 0.13 & 0.06 & 0.04 \\
        \midrule
        \multicolumn{5}{l}{\emph{point dropout} (keep $100/75/50/25\%$)} \\
        \quad dense  & 0.85 & 0.84 & 0.70 & 0.16 \\
        \quad random & 0.40 & 0.40 & 0.40 & 0.39 \\
        \midrule
        \multicolumn{5}{l}{\emph{coordinate jitter} ($\sigma{=}0/.01/.02/.05$)} \\
        \quad dense  & 0.85 & 0.84 & 0.76 & 0.25 \\
        \quad random & 0.40 & 0.40 & 0.40 & 0.32 \\
        \bottomrule
    \end{tabular}
\end{table}

\myparagraph{The dense target is what prevents collapse.} The per-point cosine correspondence target is the primary anti-collapse mechanism: removing the VICReg variance term ($\lambda{=}0$) leaves the MSR linear probe statistically unchanged (Table~\ref{tab:vicreg}; overlapping $3$-seed error bars) and far above the $0.64$ global baseline. The variance term is an inessential backstop, not the source of the gain.

\begin{table}[t]
    \centering
    \small
    \caption{\textbf{VICReg ablation} (MSR-Action3D dense linear-probe acc., mean$\pm$std over 3 seeds). Removing the variance term ($\lambda{=}0$) leaves the probe unchanged; the per-point cosine target is the primary anti-collapse mechanism.}
    \label{tab:vicreg}
    \setlength{\tabcolsep}{6pt}
    \renewcommand{\arraystretch}{1.2}
    \begin{tabular}{l|c}
        \toprule
        \textbf{Variance term} & \textbf{LP acc.} \\
        \midrule \midrule
        default ($\lambda{=}1$) & $0.836 \pm 0.019$ \\
        $\lambda{=}0$ (off)     & $0.821 \pm 0.008$ \\
        \bottomrule
    \end{tabular}
\end{table}

\myparagraph{Number of teacher cameras and target coverage.} The single-camera render leaves $\sim$$49\%$ of MSR points with a teacher target ($34\%$ on HOI4D, $16\%$ on DeformingThings4D --- its $360^\circ$ animal meshes expose one side per camera; occluded points are masked from the loss). Adding cameras raises coverage ($49{\to}74{\to}79\%$ for $1/2/4$) but \emph{not} accuracy (Table~\ref{tab:views}; flat within the seed-noise floor), so one camera suffices for single-view depth data.

\begin{table}[t]
    \centering
    \small
    \caption{\textbf{Teacher cameras vs.\ coverage and accuracy} (MSR-Action3D dense LP, mean$\pm$std over 3 seeds). This is an independent camera sweep with its own seed set, so the single-camera value ($0.828$) differs from the main result ($0.813$) within the per-seed range. Coverage rises with cameras; accuracy is flat (overlapping error bars).}
    \label{tab:views}
    \setlength{\tabcolsep}{6pt}
    \renewcommand{\arraystretch}{1.2}
    \begin{tabular}{l|cc}
        \toprule
        \textbf{\# cameras} & \textbf{coverage} & \textbf{LP acc.} \\
        \midrule \midrule
        1 (default) & $49\%$ & $0.828 \pm 0.018$ \\
        2           & $74\%$ & $0.818 \pm 0.011$ \\
        4           & $79\%$ & $0.823 \pm 0.009$ \\
        \bottomrule
    \end{tabular}
\end{table}

\myparagraph{Per-point correspondence (DeformingThings4D).} Full numbers for the per-point correspondence probe of the main paper (frozen encoder, nearest-feature matching to ground-truth mesh vertices; Table~\ref{tab:corr}). Dense $>$ random $>$ global at every frame gap, with global \emph{below} random; raw XYZ is a geometry oracle (the meshes deform slowly), so the informative comparison is among learned encoders.

\begin{table}[t]
    \centering
    \small
    \caption{\textbf{Per-point temporal correspondence on DeformingThings4D} (acc@$\tau$, $\tau{=}0.02$ of bbox diagonal; mean$\pm$std over 3 seeds). $\Delta$ is the frame gap. Raw XYZ (geometry oracle) shown for reference.}
    \label{tab:corr}
    \setlength{\tabcolsep}{6pt}
    \renewcommand{\arraystretch}{1.2}
    \begin{tabular}{l|cc}
        \toprule
        \textbf{Init} & $\Delta{=}8$ & $\Delta{=}15$ \\
        \midrule \midrule
        random        & $0.028 \pm .003$ & $0.059 \pm .023$ \\
        global        & $0.015 \pm .002$ & $0.017 \pm .003$ \\
        \textbf{dense (ours)} & $\mathbf{0.082 \pm .004}$ & $\mathbf{0.165 \pm .010}$ \\
        \midrule
        \textit{raw XYZ (oracle)} & \textit{0.152} & \textit{0.366} \\
        \bottomrule
    \end{tabular}
\end{table}

\myparagraph{Fully-supervised HOI4D segmentation ceiling.} Beyond the frozen probe of the main paper, fine-tuning the encoder end-to-end on HOI4D under full supervision (Table~\ref{tab:supceiling}) shows that dense pretraining also \emph{raises} the supervised ceiling: initializing from the dense encoder reaches $40.9$ mIoU versus $39.0$ from scratch ($+1.9$), the same ceiling-raising effect seen in action recognition.

\begin{table}[t]
    \centering
    \small
    \caption{\textbf{Fully-supervised HOI4D 4D semantic segmentation} (end-to-end fine-tuning, mIoU in \%, mean$\pm$std over 3 seeds). Dense pretraining raises the supervised ceiling over from-scratch training.}
    \label{tab:supceiling}
    \setlength{\tabcolsep}{8pt}
    \renewcommand{\arraystretch}{1.2}
    \begin{tabular}{l|c}
        \toprule
        \textbf{Initialization} & \textbf{mIoU} \\
        \midrule \midrule
        from scratch          & $39.0 \pm 0.5$ \\
        \textbf{dense (ours)} & $\mathbf{40.9 \pm 0.3}$ \\
        \bottomrule
    \end{tabular}
\end{table}

\section{Detailed ``what is not the lever'' decompositions}
\label{supp:ablations}

This section reports the detailed numbers behind the main paper's analysis of what is \emph{not} the lever: neither teacher modality nor the latent-forecasting (world-model) components is the mechanism that drives cross-modal transfer to 4D point clouds --- the per-point correspondence is.

\myparagraph{Teacher modality (global supervision).} With supervision reduced to a single global embedding per clip, a video teacher (V-JEPA\,2) and an image teacher (DINOv2) give statistically indistinguishable linear-probe accuracy --- and this tie holds on \emph{both} backbones we tested (\textsc{Point4D} in the main paper and \textsc{P4Transformer} here; Table~\ref{tab:app-modality}), so it is not a backbone artifact. The dense formulation instead reaches $0.81$ with the image teacher, so the gain comes from \emph{granularity}, not modality.

\begin{table}[t]
    \centering
    \small
    \caption{Teacher modality under \emph{global} supervision (linear-probe acc.) on \emph{both} backbones. The two teachers are tied within the per-seed spread on both backbones; cf.\ dense distillation at $0.81$/$0.88$ (image/video) in the main paper.}
    \label{tab:app-modality}
    \setlength{\tabcolsep}{6pt}
    \renewcommand{\arraystretch}{1.2}
    \begin{tabular}{l|cc}
        \toprule
        \textbf{Global teacher} & \textbf{\textsc{Point4D}} & \textbf{\textsc{P4Transformer}} \\
        \midrule \midrule
        V-JEPA\,2 (video) & 0.61 & 0.800 \\
        DINOv2 (image)    & 0.64 & 0.804 \\
        \bottomrule
    \end{tabular}
\end{table}

\begin{figure*}[t]
  \centering
  \includegraphics[width=0.8\linewidth]{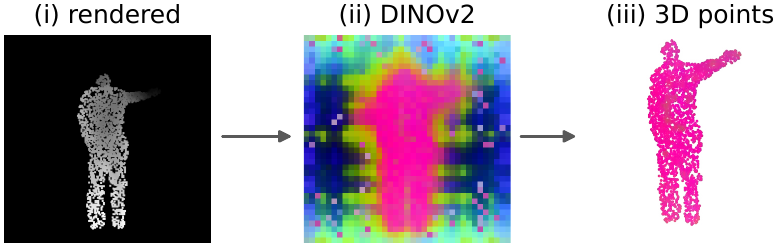}
  \caption{\textbf{The dense distillation pipeline} on one MSR-Action3D frame: the input point cloud is (i)~rendered to a depth-shaded image and fed to DINOv2, giving (ii)~DINOv2 patch features (PCA$\to$RGB; the subject separates cleanly from the background), which are (iii)~back-projected onto the 3D points via the render correspondence. (ii) and (iii) share one PCA color space, so the subject's colors match in both.}
  \label{fig:supp-pipeline}
\end{figure*}

\myparagraph{Latent-forecasting components (global regime).} Table~\ref{tab:app-worldmodel} decomposes the world-model variants we explored. These are \emph{single-seed} runs on a lightweight voxel encoder (\textsc{Simple4D}) in the global-target regime; we include them for completeness and \emph{do not} build the method on them. Two points stand out. First, the \emph{entire} regime tops out at $0.33$ --- well below dense distillation's $0.81$ --- so no arrangement of these components is competitive with dense correspondence. Second, on the stronger \textsc{P4Transformer} backbone with multiple seeds, the pairwise differences among these variants fell within the $\sim$0.04 single-seed noise floor, i.e.\ they are not robust. We therefore treat the dense correspondence, not the forecasting machinery, as the contribution.

\begin{table}[t]
    \centering
    \small
    \caption{Exploratory decomposition of the latent-forecasting (world-model) components (\emph{single seed}, lightweight \textsc{Simple4D}, V-JEPA\,2 global target, MSR-Action3D linear-probe acc.). The whole regime is far below dense distillation ($0.81$); on \textsc{P4Transformer} with 3 seeds, the differences are within the noise floor. Reported for completeness only.}
    \label{tab:app-worldmodel}
    \setlength{\tabcolsep}{6pt}
    \renewcommand{\arraystretch}{1.2}
    \begin{tabular}{l|c}
        \toprule
        \textbf{Variant} & \textbf{LP acc.} \\
        \midrule \midrule
        full (field $+$ rollout $+$ consistency) & 0.331 \\
        \quad $-$ consistency (field $+$ rollout) & 0.258 \\
        direct predictor (no field)              & 0.222 \\
        field, no rollout                        & 0.171 \\
        co-temporal ($\Delta{=}0$)               & 0.160 \\
        \bottomrule
    \end{tabular}
\end{table}

\section{Dense distillation on \textsc{P4Transformer}}
\label{supp:p4t}

We applied the same dense correspondence distillation to the official \textsc{P4Transformer} backbone (its CUDA point-4D-convolution ops). Across four variants, the linear probe lands near $0.60$ --- far below \textsc{Point4D}'s $0.813$ under the identical objective, teacher, and budget (Table~\ref{tab:app-p4t}). A final LayerNorm fixes a feature-magnitude explosion (the scale-invariant cosine loss otherwise lets the unnormalized transformer output drift to standard deviation $\sim$80) but does not move the probe; eval-time feature standardization only reaches $0.62$; and a less-pooled (finer-token) configuration does not help either. This is most consistent with the backbone's pooled tokenization --- designed to funnel a clip toward a single classification token --- which leaves little per-point structure for the dense objective, rather than with optimization or feature scale (cf.\ the main paper's discussion). We keep the dependency-free \textsc{Point4D} as the main backbone for this reason.

\begin{table}[t]
    \centering
    \small
    \caption{Dense distillation across backbones (MSR-Action3D linear-probe acc.). The official \textsc{P4Transformer}'s pooling discards per-point structure, so the same dense objective underperforms \textsc{Point4D} regardless of normalization or token granularity. 3-seed mean$\pm$std where available.}
    \label{tab:app-p4t}
    \setlength{\tabcolsep}{6pt}
    \renewcommand{\arraystretch}{1.2}
    \begin{tabular}{l|c}
        \toprule
        \textbf{Backbone (dense distillation)} & \textbf{LP acc.} \\
        \midrule \midrule
        \textbf{Point4D (main)}                 & \textbf{0.813 $\pm$ 0.021} \\
        \midrule
        P4Transformer, official pooling         & $\sim$0.59 \\
        \quad $+$ final LayerNorm               & 0.600 $\pm$ 0.031 \\
        \quad $+$ finer tokens (less pooling)   & 0.607 $\pm$ 0.022 \\
        \bottomrule
    \end{tabular}
\end{table}

\section{Additional qualitative visualizations}
\label{supp:viz}

Each map colors every input point by a $3$-component PCA (mapped to RGB) of its nearest encoder token's feature --- the same rule used for the main paper's Fig.~2 --- with the encoder frozen.

\begin{figure*}[t]
  \centering
  \includegraphics[width=0.9\linewidth]{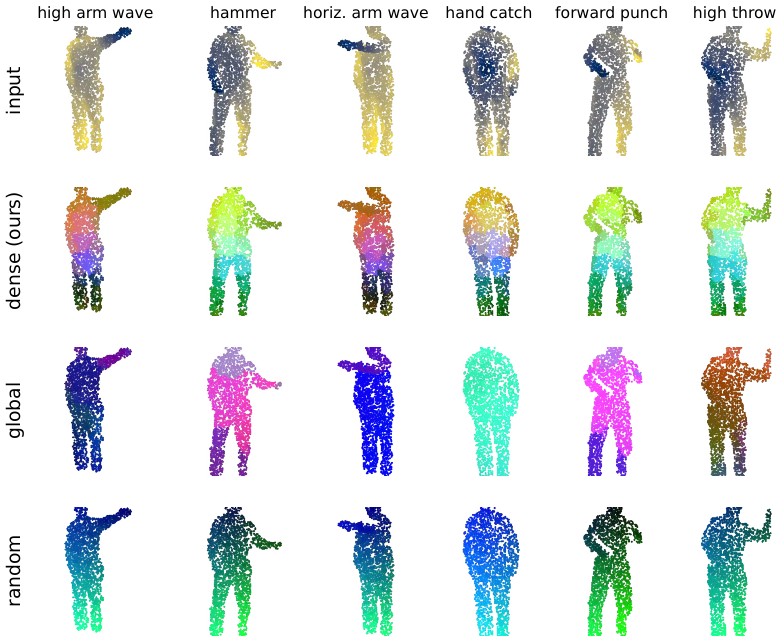}
  \caption{\textbf{Dense vs.\ global vs.\ random per-point features} (MSR-Action3D, six actions). Extended version of the main paper figure: each input point is colored by the PCA-to-RGB of its nearest encoder token. The \emph{dense} encoder forms consistent body-part regions (distinct torso, legs, arms, head) repeated across actions; the \emph{global}-distilled encoder collapses toward a blobbier, near-uniform per-clip coloring; and random init is unstructured --- a qualitative view of why global distillation discards the per-point structure a dense task needs (cf.\ the HOI4D segmentation result in the main paper). \textbf{Rows} are encoders, \textbf{columns} are actions; each encoder row has its own PCA color space (fit over all clips), so hues are comparable across actions \emph{within} a row but not across the encoder rows --- the comparison is spatial \emph{coherence}.}
  \label{fig:supp-densevsglobal}
\end{figure*}

{
    \small
    \bibliographystyle{ieeenat_fullname}
    \bibliography{refs}
}

\end{document}